\documentclass[sigconf]{acmart}
\AtBeginDocument{%
  }

\copyrightyear{2026}
\acmYear{2026}
\setcopyright{cc}
\setcctype{by}
\acmConference[MM '26]{Proceedings of the 34th ACM International Conference on Multimedia}{November 10--14, 2026}{Rio de Janeiro, Brazil}
\acmBooktitle{Proceedings of the 34th ACM International Conference on Multimedia (MM '26), November 10--14, 2026, Rio de Janeiro, Brazil}
\acmDOI{10.1145/3767308.3835745}
\acmISBN{979-8-4007-2213-4/2026/11}

\usepackage[ruled,linesnumbered]{algorithm2e}
\usepackage{amsmath, amsthm, bm, mathtools}
\usepackage{graphicx}
\usepackage{float}
\usepackage{tikz}
\usepackage{adjustbox} 
\usepackage[table,xcdraw]{xcolor}
\usepackage{colortbl}
\usepackage{booktabs}
\usepackage{multirow}
\usepackage{microtype}
\usepackage{soul}   
\usepackage{ragged2e}  
\usepackage{tabularx}
\usepackage{makecell}

\begin{document}

\title{More Perspectives, Stronger Signals: Multi-Perspective Enhancement and Progressive Fusion for Multimodal Entity Representation Learning}

\author{Chenyi Xiong}
\affiliation{%
  \institution{School of Computer Science, Hubei University}
  \city{Wuhan}
  \country{China}
}
\email{xiongchenyi@stu.hubu.edu.cn}

\author{Yan Zhang}
\affiliation{%
  \institution{School of Computer Science, Hubei University}
  \city{Wuhan}
  \country{China}}
\email{zhangyan@hubu.edu.cn}
  
\author{Jing Hu}
\affiliation{%
  \institution{School of Cyber Science and Technology, Hubei University}
  \city{Wuhan}
  \country{China}
}
\email{jinghu2024@stu.hubu.edu.cn}

\author{Ziyue Qin}
\affiliation{%
  \institution{School of Computer Science, Hubei University}
  \city{Wuhan}
  \country{China}
}
\email{qinziyue@stu.hubu.edu.cn}

\author{Kui Xiao}
\affiliation{%
  \institution{School of Computer Science, Hubei University}
  \city{Wuhan}
  \country{China}
  }
  \email{xiaokui@hubu.edu.cn}

\author{Xiaopan Lyu}
\affiliation{%
  \institution{School of Computer Science, Hubei University}
  \city{Wuhan}
  \country{China}
  }
  \email{xiaopan.lyu@hubu.edu.cn}

\author{Xiaoju Hou}
\affiliation{%
  \institution{Institute of Vocational Education, Guangdong Industry Polytechnic University}
  \city{Guangzhou}
  \country{China}}
  \email{2023030010@gdip.edu.cn}

\author{Zhifei Li}
\authornote{Corresponding author.}
\affiliation{%
  \institution{School of Computer Science, Hubei Key Laboratory of Big Data Intelligent Analysis and Application, Hubei University}
  \city{Wuhan}
  \country{China}
  }
\email{zhifei1993@hubu.edu.cn}

\renewcommand{\shortauthors}{Chenyi Xiong et al.}
\begin{abstract}
Learning effective multimodal entity representations is fundamental for reasoning tasks such as multimodal knowledge graph completion (MMKGC). However, existing methods often suffer from semantic over-smoothing within modalities and ineffective noise filtration across modalities, particularly under sparse or ambiguous conditions. To overcome these limitations, we propose PrismF, a unified framework that synergizes multi-perspective enhancement with progressive fusion to extract stronger signals from diverse inputs. PrismF enhances fine-grained intra-modal semantics through a multi-perspective mechanism that decomposes each modality into complementary views and constrains them with a decoupling loss to reduce representation collapse. Furthermore, it improves cross-modal integration through a progressive fusion strategy that dynamically calibrates inter-modal interactions, enabling the model to emphasize informative signals while suppressing noisy or unreliable ones. Extensive experiments on three public benchmarks show that PrismF
achieves the strongest overall performance, including relative
improvements of 4.04\% in MRR and 11.17\% in Hits@1 on KVC16K. Our code can be found at \url{https://github.com/HubuKG/PrismF}. 
\end{abstract}



\maketitle

\section{Introduction}
Multimodal Knowledge Graphs (MMKGs) \citep{1} extend knowledge graphs with heterogeneous entity signals, supporting applications such as large language models \citep{LLM,6}, recommendation systems \citep{3,recom}, and computer vision \citep{4,5}. However, incompleteness and sparse connectivity hinder reliable inference \citep{neg,8}, making multimodal knowledge graph completion (MMKGC) essential.
\begin{figure}[t]
    \centering
    \includegraphics[width=1\linewidth]{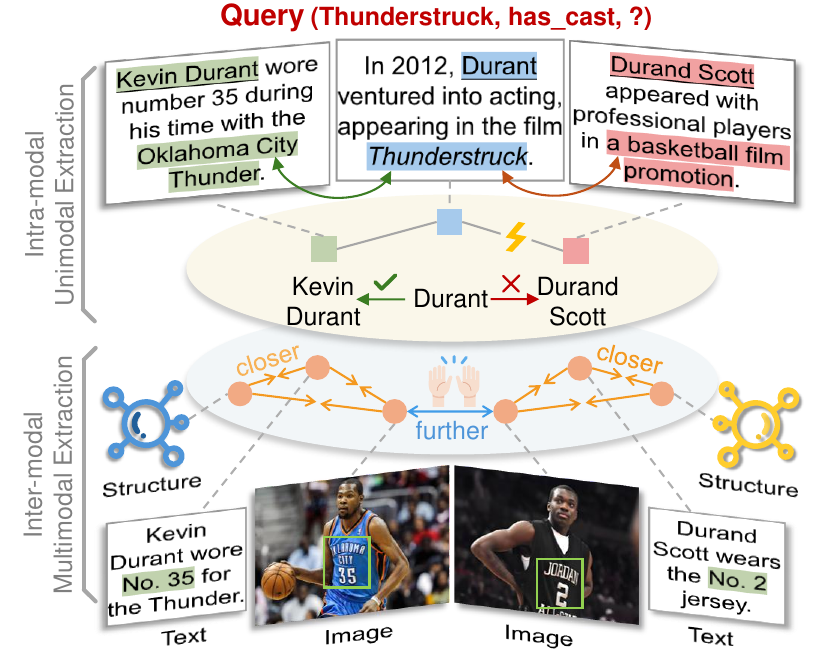}
    \caption{Illustration of reasoning challenges in MMKGC. When querying (\textit{Thunderstruck, has\_cast, ?}), textual ambiguity and visual similarity between entities with similar names (\textit{Kevin Durant} vs. \textit{Durand Scott}) lead to reasoning confusion.}
    \label{fig1}
\end{figure}

Most MMKGC methods \citep{9,10,LBMKGC} adopt an encode-and-fuse paradigm, where modality features are first extracted by pre-trained models and then combined for link prediction. Although effective, this paradigm does not fully match the demands of MMKGC. The key challenge is not merely incorporating more modalities, but preserving discriminative cues within each modality and integrating heterogeneous signals according to their contextual reliability.

This mismatch stems from two tightly coupled challenges. First, pre-trained models favor coarse semantic abstraction, potentially suppressing the subtle textual and visual cues required to distinguish semantically similar entities. As illustrated in Figure~\ref{fig1}, this leads to feature homogenization, where the fine-grained cues such as the jersey numbers or facial characteristics that distinguish "\textit{Kevin Durant}" from "\textit{Durand Scott}", may be overshadowed by shared high-level semantics such as "basketball player". Second, multimodal signals vary substantially in quality across relations and reasoning instances. Noise, ambiguity, and incompleteness may cause conventional fusion mechanisms to rely excessively on a dominant yet unreliable modality, resulting in modality collapse \cite{collapse}. These challenges reinforce each other. Once discriminative cues are weakened during encoding, downstream fusion can only combine already diluted representations. Conversely, even informative modality-specific representations may still impair reasoning if their contributions are not calibrated to the current context.

To address these, we propose PrismF, a unified framework for MMKGC. The core idea is to treat multimodal reasoning as a two-stage process that first recovers fine-grained discriminative cues and then adaptively calibrates cross-modal interactions. First, Multi-Perspective Enhancement (MuPE) expands each modality into multiple complementary perspectives and encourages their diversity through a decoupling objective, thereby preserving distinctions that may be lost in a single smoothed representation. Second, Progressive Modality Fusion (PMF) builds on these enriched features to estimate modality confidence under structural guidance and progressively adjust the contribution of each modality during fusion. By strengthening informative evidence while suppressing unreliable signals, PrismF jointly mitigates feature homogenization and noise propagation. Extensive experiments on three public datasets demonstrate the effectiveness of PrismF. 

Our main contributions are summarized as follows:
\begin{itemize}
\item We identify a core challenge in MMKGC arising from the joint need to preserve fine-grained discriminative cues within modalities and to calibrate multimodal fusion according to contextual reliability.
\item We propose PrismF, a unified framework that addresses this challenge by coupling Multi-Perspective Enhancement for complementary intra-modal enrichment with Progressive Modality Fusion for inter-modal calibration.
\item We conduct extensive experiments on DB15K, MKG-Y, and KVC16K, and the results show that PrismF consistently outperforms strong baselines across multiple metrics.
\end{itemize}

\section{Related Work}
\subsection{Knowledge Graph Completion}
Knowledge Graph Completion (KGC) \citep{KGsurvey,KGCsurvey} aims to predict missing triples by modeling the structural dependencies among entities and relations. Early embedding-based methods \citep{KGE1,KGE2} learn low-dimensional representations of entities and relations and assess triple plausibility with predefined scoring functions. Translation-based models, such as TransE \citep{TransE} and RotatE \citep{RotatE}, represent relations as geometric transformations in the embedding space. In contrast, factorization-based models, including DistMult \citep{Distmult}, ComplEx \citep{CompIEX}, and TuckER \citep{Tucker}, model semantic interactions through bilinear or tensor decomposition forms. Other methods, such as ConvE \citep{ConvE} and M-DCN \citep{M-DCN}, further enhance representation learning with convolutional architectures. More recently, GNN-based methods \citep{12,15} have attracted increasing attention for their ability to exploit graph structure and neighborhood dependencies. 

Although these methods have achieved strong performance, they mainly rely on structural information and often overlook the rich multimodal signals associated with real-world entities. This limitation has motivated the development of MMKGC methods that incorporate textual and visual information into KGC.

\subsection{Multimodal Knowledge Graph Completion}
MMKGC extends traditional KGC \citep{7,incom,multi} by jointly modeling structural triples with auxiliary modalities such as textual descriptions and visual content. Recent MMKGC methods \citep{16,align,18} have increasingly focused on improving cross-modal interaction and multimodal fusion. Some methods design more refined alignment mechanisms, such as optimal transport-based approaches \citep{OTKGE,22,20}, to reduce distributional discrepancies across modalities. Other methods explore dynamic fusion strategies, including modality ensemble approaches \citep{IMF,MoSE} and adaptive fusion frameworks \citep{MYGO,APKGC,SNAG,HFR-MKGC}, to combine modality-specific cues for prediction. In addition, adversarial learning methods \citep{native,ADAMF} have been introduced to improve modality-invariant representation learning and generalization. Beyond representation and fusion, several studies \citep{vbkgc,MANS,neg} further enhance MMKGC by refining the training process.

Different from prior methods that primarily focus on cross-modal alignment or adaptive fusion at the final integration stage, PrismF formulates the task as two coupled processes: intra-modal perspective enrichment and inter-modal reliability calibration. This design enables the model to preserve fine-grained modality-specific evidence before fusion and to perform reliability-aware multimodal integration on top of these enriched representations.

\begin{figure*}
    \centering
    \includegraphics[width=0.95\linewidth]{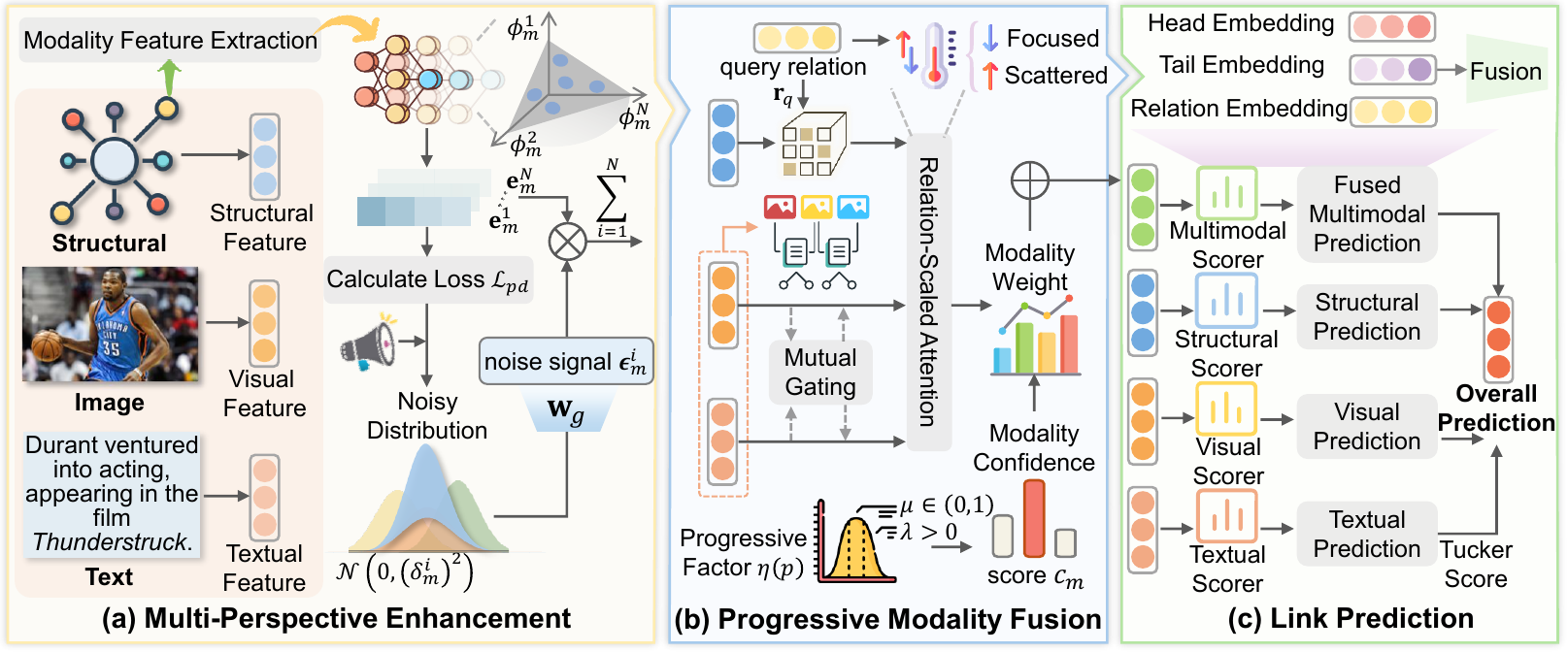}
    \caption{Overview of the PrismF framework. (a) Multi-Perspective Enhancement: decomposes each modality into multiple learnable perspectives with noise modeling to capture intra-modal variability and enhance fine-grained representations. (b) Progressive Modality Fusion: dynamically integrates cross-modal features through confidence calibration, mutual gating, and relation-scaled attention to achieve multimodal fusion. (c) Link Prediction: leverages fused and modality-specific representations with Tucker scoring for robust link prediction in KGs.}
    \label{f2}
\end{figure*}

\section{Methodology}

In this section, we present PrismF, a unified framework designed to improve reasoning over heterogeneous modality signals. As illustrated in Figure~\ref{f2}, PrismF consists of three key components: Multi-Perspective Enhancement, Progressive Modality Fusion, and Link Prediction, which progressively enhance modality-specific representations, integrate multimodal information, and perform joint prediction in MMKGs, respectively.

\subsection{Task Definition}

A KG can be typically represented as $\mathcal{G}=(\mathcal{E},\mathcal{R},\mathcal{T})$, where $\mathcal{E}$ is the entity set, $\mathcal{R}$ is the relation set, and $\mathcal{T}=\left\{\left(h,r,t\right)|h,t\in\mathcal{E},r\in\mathcal{R}\right\}$ is the triple set. To characterize MMKGs, we introduce a modality set $\mathcal{M}=\{t,v,s\}$, corresponding to textual, visual, and structural modalities. For each entity $e\in\mathcal{E}$, $X_m(e)$ denotes its modality-specific information under modality $m\in\mathcal{M}$, while the triple set $\mathcal{T}$ is regarded as the structural modality $s$.

MMKGC aims to learn a scoring function $S(h,r,t): \mathcal{E} \times \mathcal{R} \times \mathcal{E} \rightarrow \mathbb{R}$ that evaluates triple plausibility using fused modality embeddings $\mathbf{e}_m$ ($m \in \mathcal{M}$). Negative sampling generates negative triples by replacing heads or tails of positive triples for contrastive training. During reasoning, MMKGC is evaluated via link prediction by ranking candidate triples for queries like $\left(?,r,t\right)$ or $\left(h,r,?\right)$.

\subsection{Multi-Perspective Enhancement}
Traditional multimodal representation approaches often rely on single-view embeddings, which fail to capture the intricate and nuanced information within each modality. For example, in image and text, aspects such as color, spatial relationships, or contextual semantics may exert different influences depending on the task. To address this, we propose a Multi-Perspective Enhancement (MuPE) mechanism, which introduces multiple learnable perspectives to decompose and represent fine-grained intra-modal features. 

Specifically, we use pre-trained models such as VGG \citep{VGG} for image features and BERT \citep{BERT} for text features to extract raw embeddings $\mathbf{e}_m$, while structural modality features are directly learned from triple data during training. For each modality $m\in\left\{t,v,s\right\}$, MuPE constructs $N$ learnable perspective functions $P_m=\left\{\phi_m^1,\phi_m^2,\cdots,\phi_m^N\right\}$, each designed to transform the input embedding $\mathbf{e}_m$ into perspective-specific feature spaces. The feature representation for the $i$-th perspective is computed as:
\begin{equation}
\mathbf{h}_m^i=\phi_m^i\left(\mathbf{e}_m\right)=\mathbf{W}_m^i\cdot Dropout\left(\mathbf{e}_m-\mathbf{b}_m\right), \quad i=1,\dots,N,
\end{equation}
\begin{equation}
\mathbf{e}_m^{i}=\text{LayerNorm}(\mathbf{h}_m^{i}),
\end{equation}
where $\mathbf{W}_m^i$ is a learnable projection matrix, $\mathbf{b}_m$ is a learnable bias term, and the dropout operation mitigates overfitting by reducing reliance on specific input dimensions.

To capture uncertainty in perspective-specific representations and improve robustness to unstable fine-grained cues, we further introduce a stochastic refinement module. Given a normalized perspective representation $\mathbf{e}_m^{i}$ (for modality $m$ and perspective $i$), we add a Gaussian noise term $\boldsymbol{\epsilon}_m^i$:
\begin{equation}
\delta_m^i = \text{Softplus}(\mathbf{W}_n^i \cdot \mathbf{e}_m^i) \odot \psi(\mathbf{e}_m^i),
\end{equation}
\begin{equation}
\boldsymbol{\epsilon}_m^i \sim \mathcal{N} \left( \mathbf{0}, \left( \delta_m^i \right)^2 \right),
\end{equation}
where $\mathbf{W}_n^i$ is a learnable projection matrix, $\psi(\cdot)$ is a two-layer MLP with GELU activation, and the Softplus function ensures non-negativity and smoothness. This stochastic refinement injects mild perturbations during training and discourages over-reliance on unstable perspective-specific cues.

Perspective-specific embeddings are aggregated according to their relation-specific importance. The attention weight $g_m^i$ for the $i$-th perspective is computed as:
\begin{equation}
g_m^i = \frac{\exp\left( \frac{ {\mathbf{w}_g^i}^{\top}(\mathbf{e}_m^i + \boldsymbol{\epsilon}_m^i) }{ \sigma(\mathbf{r}_q) } \right)}
{\sum_{j=1}^{N} \exp\left( \frac{ {\mathbf{w}_g^j}^{\top}(\mathbf{e}_m^j + \boldsymbol{\epsilon}_m^j) }{ \sigma(\mathbf{r}_q) } \right)},
\end{equation}
where $\mathbf{w}_g^i$ is a learnable parameter vector, and $\sigma(\mathbf{r}_q) = \text{Softplus}(\mathbf{W}_{t} \cdot {\mathbf{r}}_{q} + \mathbf{b}_{t})$ is a relation-aware normalization factor that calibrates perspective weights under different query relations. The final aggregated embedding $\widetilde{\mathbf{e}}_m$ is defined as:
\begin{equation}
\widetilde{\mathbf{e}}_m=\sum_{i=1}^{N} g_m^i \mathbf{e}_m^i.
\end{equation}

To ensure that the learned perspectives capture complementary rather than redundant information, we introduce a perspective decoupling loss $\mathcal{L}_{pd}$. This loss minimizes the soft orthogonality constraint on the perspective embeddings. For a batch of size $B$, let $E_{m,k} = [{\mathbf{e}}_{m,k}^1, \ldots, {\mathbf{e}}_{m,k}^N]$ denote the matrix of perspective embeddings for the $k$-th sample. We compute the Gram matrix $G_{m,k} = E_{m,k}^\top E_{m,k}$ and penalize deviations from orthogonality:
\begin{equation}
\mathcal{L}_{pd} = \frac{1}{|\mathcal{M}|\, B} 
\sum_{m \in \mathcal{M}} \sum_{k=1}^{B}
\frac{ \| G_{m,k} - I \|_{F}^{2} }{ N (N-1) },
\end{equation}
where $\mathcal{M}$ is the set of modalities and
$I = I_{N} \in \mathbb{R}^{N \times N}$ denotes the identity matrix. This term drives off-diagonal entries of $G_{m,k}$ toward zero to reduce inter-perspective inner-product correlation while keeping the diagonal near one, thereby encouraging each perspective to capture distinct information without inflating vector norms.

MuPE enables the model to selectively focus on the most informative perspectives for the target relation while maintaining diversity and robustness across embeddings. By enriching each modality’s representation with complementary perspectives and reducing intra-modal similarity, MuPE enhances the model’s ability to capture complex dependencies and fine-grained semantic cues.

\begin{table}[t]
\centering
\caption{Statistics of the multimodal knowledge graph datasets in our experiments. The image and text modality features are provided by the original datasets and are kept the same for all baselines.}
\renewcommand\arraystretch{1.1}
\tabcolsep=0.2cm
\resizebox{\columnwidth}{!}{%
\begin{tabular}{lccccccccc}
\toprule
\multirow{2}{*}{\textbf{Dataset}} & 
\multirow{2}{*}{\textbf{\#Entity}} & 
\multirow{2}{*}{\textbf{\#Relation}} & 
\multirow{2}{*}{\textbf{\#Train}} & 
\multirow{2}{*}{\textbf{\#Valid}} & 
\multirow{2}{*}{\textbf{\#Test}} & 
\multicolumn{2}{c}{\textbf{Image}} & 
\multicolumn{2}{c}{\textbf{Text}} \\
\cmidrule(lr){7-8} \cmidrule(lr){9-10}
 &  &  &  &  &  & Num. & Dim. & Num. & Dim. \\
\midrule

\textbf{DB15K}  & 12,842 & 279 & 79,222  & 9,902  & 9,904  & 12,818 & 4,096 & 9,078  & 768 \\
\textbf{MKG-Y}   & 15,000               & 28                    & 21,310            & 2,665             & 2,663            & 14,244 & 383  & 12,305 & 384 \\
\textbf{KVC16K}  & 16,015               & 4                     & 180,190           & 22,523            & 22,525           & 14,822 & 768  & 14,822 & 768 \\
\bottomrule
\end{tabular}}
\label{tab1}
\end{table}

\subsection{Progressive Modality Fusion}
In MMKGC tasks, different modalities vary in richness and reliability depending on the sample and relational context. Prior work has shown that static fusion often yields modality dominance or collapse, where a noisy or superficially salient modality overwhelms others, causing (i) amplification of spurious signals and (ii) suppression of complementary but weaker modalities \citep{native,MoSE}. Motivated by these observations, we design a Progressive Modality Fusion (PMF) mechanism to dynamically recalibrate modality contributions and incrementally combine modality-specific representations according to confidence scores and relation-aware signals. This design ensures that the fusion process remains dynamic and interpretable across different training stages and relational contexts.

\subsubsection{Reliability Calibration}
We quantify the contribution of each modality by estimating a query-specific confidence score for modality $m\in\{t,v,s\}$ under query $q$. Using the modality-enhanced embedding ${\widetilde{\mathbf{e}}}_{m,q}$ from the MuPE module, we first compute a confidence logit:
\begin{equation}
s_{m,q}=\mathbf{w}_{conf}^{m\top}\,\mathrm{GELU}(\mathbf{W}_s\cdot {\widetilde{\mathbf{e}}}_{m,q}+\mathbf{b}_s),
\end{equation}
where $\mathbf{w}_{conf}^m$ is a modality-specific confidence vector, and $\mathbf{W}_s,\mathbf{b}_s$ are shared learnable parameters. We then normalize the logits across modalities for the same query:
\begin{equation}
c_{m,q}=\frac{\exp(s_{m,q})}{\sum_{m^\prime\in\{s,v,t\}}\exp(s_{m^\prime,q})}.
\end{equation}
In this way, $c_{m,q}$ reflects the relative reliability of modality $m$ conditioned on the current query context.

To guide the fusion process throughout training, we incorporate a progressive gating function that adaptively modulates confidence scores according to normalized training progress:
\begin{equation}
\eta\left(p\right)=\frac{1}{1+e^{-\lambda\left(p-\mu\right)}},
\label{eq9}
\end{equation}
where $p\in [0,1]$ denotes the normalized training progress calculated as the ratio between the current training step and the total number of training steps, while $\lambda$ and $\mu$ are hyperparameters controlling the steepness and center of the gating curve. The final calibrated confidence score is defined as:
\begin{equation}
\widetilde{c}_{m,q} = c_{m,q} \cdot \eta(p).
\end{equation}

\subsubsection{Mutual Gating Strategy}
To facilitate semantic alignment and filter redundant information, we introduce a mutual gating strategy between the visual and textual modalities. Specifically, visual features are modulated by text embeddings, and vice versa:
\begin{equation}
\mathbf{g}_{v \leftarrow t} = \sigma(\mathbf{W}_{vt} \cdot \widetilde{\mathbf{e}}_{t,q} + \mathbf{b}_{vt}), \quad 
\mathbf{g}_{t \leftarrow v} = \sigma(\mathbf{W}_{tv} \cdot \widetilde{\mathbf{e}}_{v,q} + \mathbf{b}_{tv}),
\end{equation}
\begin{equation}
\widehat{\mathbf{e}}_{v,q} = \widetilde{\mathbf{e}}_{v,q} \odot \mathbf{g}_{v \leftarrow t}, \quad 
\widehat{\mathbf{e}}_{t,q} = \widetilde{\mathbf{e}}_{t,q} \odot \mathbf{g}_{t \leftarrow v},
\end{equation}
where $\odot$ is the point-wise operator. 
For the structural modality, which offers stable graph-level signals but limited semantic detail, we introduce a relation-aware projection to enhance its alignment with the query relation. The final representations are defined as follows:
\begin{equation}
\mathbf{e}_{m,q}^{\text{final}} =
\begin{cases}
\widehat{\mathbf{e}}_{m,q}, & \text{if } m \in \{v, t\}, \\
\left( \mathbf{W}_r \cdot \mathbf{r}_q + \mathbf{b}_r \right) \odot \widetilde{\mathbf{e}}_{s,q}, & \text{if } m = s,
\end{cases}
\end{equation}
where $\mathbf{W}_r$ is a learnable weight matrix, $\mathbf{b}_r$ is a bias term. This design ensures effective cross-modal alignment while leveraging structural signals for enhanced relational reasoning, improving their discriminative utility.

\begin{table*}[h]
\centering
\caption{Performance comparison of different knowledge graph completion models. The optimal results are highlighted in \textbf{bold}, while the second optimal results are \underline{underlined}.}
\definecolor{GroupA}{gray}{0.93}
\definecolor{GroupB}{HTML}{F5E7EA}
\definecolor{GroupC}{HTML}{EEF4FC}
\definecolor{GroupD}{HTML}{F8F6EC}

\renewcommand\arraystretch{0.9}
\resizebox{\textwidth}{!}{\begin{tabular}{lccccccccccccccccc}
\toprule
\multirow{3}{*}{\textbf{Model}} & \multicolumn{4}{c}{\textbf{DB15K}}                                  &           & \multicolumn{4}{c}{\textbf{MKG-Y}}                             &           & \multicolumn{4}{c}{\textbf{KVC16K}}                                  \\ \cmidrule{2-5} \cmidrule{7-10} \cmidrule{12-15} 
                                 & \multirow{2}{*}{MRR} & \multicolumn{3}{c}{Hits}                         &           & \multirow{2}{*}{MRR} & \multicolumn{3}{c}{Hits}                         &           & \multirow{2}{*}{MRR} & \multicolumn{3}{c}{Hits}                         \\ \cline{3-5} \cline{8-10} \cline{13-15} 
                                 &                     & @1             & @3             & @10            &           &                     & @1             & @3             & @10            &           &                     & @1             & @3             & @10            \\ \midrule

\rowcolor{GroupA}
\multicolumn{15}{c}{\textit{Unimodal KGC Models}}                                                                                                                                                                                                                                            \\ \midrule
\rowcolor{GroupD} TransE (NeurIPS'13)                           & 24.86                 & 12.78          & 31.48          & 47.07                    &           & 30.73                & 23.45          & 35.18          & 43.37 &           & 8.54                 & 0.64          & 10.97          & 23.42         \\
\rowcolor{GroupD} DistMult (ICLR'15)                        & 23.03                 & 14.78          & 26.28          & 39.59          &           &  25.04                & 19.33          & 27.80          & 35.95          &           & 6.37                & 3.03          & 6.11          & 12.61          \\
\rowcolor{GroupD} ComplEx (ICML'16)                         & 27.48                   & 18.37          & 31.57          & 45.37          &           & 28.71 &22.26 &32.12 &40.93
                   &           & 12.85                & 7.48          & 13.79          & 23.18          \\
\rowcolor{GroupD} RotatE (ICLR'19)  
& 29.28                 & 17.87          & 36.12          & 49.66          &           & 34.95 &29.10 &38.35 &45.30         &           & 14.33                & 8.25          & 15.37          & 26.17          \\
 \midrule
\rowcolor{GroupA}\multicolumn{15}{c}{\textit{Multimodal KGC Models}}                                                                                                                                                                                                                                          \\ \midrule
\rowcolor{GroupC}IKRL (IJCAI'17)                             & 26.82                 & 14.09          & 34.93          & 49.09          &           & 33.22 &30.37 &34.28 &38.26          &           & 11.11                & 5.42          & 11.46          & 22.39          \\
\rowcolor{GroupC}TransAE (IJCNN'19)                          & 28.09                 & 21.25          & 31.17          & 41.17          &           & 28.10 &25.31 &29.10 &33.03          &           & 10.81                & 5.31          & 11.34          & 21.89          \\
\rowcolor{GroupC}RSME (ACM MM'21)                             & 29.76                 & 24.15          & 32.12          & 40.29         &           & 34.44 &31.78 &36.07 &39.09          &           & 12.31                & 7.14          & 13.21          & 22.05          \\
\rowcolor{GroupC}VBKGC (arXiv'22)                            & 30.61                  & 19.75             & 37.18              & 49.44              &           & 37.04 &33.76 &38.75 &42.30          &           & 14.66                & 8.28          & 15.81          & 27.04          \\
\rowcolor{GroupC}OTKGE (NeurIPS'22)                           & 23.86                   & 18.45             & 25.89              & 34.23              &           & 35.51 &31.97 &37.18 &41.38
          &           & 8.77                & 5.01          & 9.31          &15.55          \\
\rowcolor{GroupC}IMF (WWW'23)                        & 32.25            & 24.20    & 36.00    & 48.19    &     & 35.79 & 32.95 & 37.14 & 40.63         &           & 12.01           & 7.42    & 12.82   & 21.01    \\
\rowcolor{GroupC}QEB (ACM MM'23)                             & 28.18                 & 14.82          & 36.67          & 51.55          &           & 34.37 & 29.49 & 36.95 & 42.32
        &           & 12.06                & 5.57          & 13.03          & 25.01          \\
\rowcolor{GroupC}MANS (IJCNN'23)                          & 28.82                 & 16.97          & 36.58          & 49.26          &           & 29.03 & 25.25 & 31.35 & 34.49
   &           & 10.42                & 5.21          & 11.01          & 20.45          \\
\rowcolor{GroupC}MMRNS (ACM MM'22)                          & 32.68                 & 23.01          & 37.86          & 51.01          &           & 35.93  &30.53  &39.07  &45.47    &           & 13.31                & 7.51          & 14.19          & 24.68         \\
\rowcolor{GroupC}AdaMF (COLING'24)                          & 32.51                 & 21.31          & 39.67          & 51.68          &           & 38.06 & 33.49 & 39.44 & 45.48    &           & 15.26                & 8.56          & 16.71          & 28.29         \\
\rowcolor{GroupC}MYGO (AAAI'25)                          & \underline{37.72}                 & \underline{30.08}          & 41.26          & 52.21          &           & 38.24                   & \underline{35.01}          & 39.84              &   44.19        &                &\underline{16.32}          & \underline{9.58}          & \underline{17.26}     &\underline{28.57}   \\
\rowcolor{GroupC}LBMKGC (NeurIPS'25)                          & 37.23                 & 27.78          & 42.75          & 54.71          &          &  \underline{40.03}                   &  33.89        & \underline{43.11}         & \textbf{50.81}    &           & -              & -         & -          & -       \\
\rowcolor{GroupC}HFR-MKGC (AAAI'26)                          & 35.20                 & 23.04          & \underline{43.25}          & \underline{56.41}          &          &  39.00                   &  34.41        & 42.12         & \underline{47.31}    &           & -              & -         & -          & -       \\\midrule
\rowcolor{GroupB}\textbf{PrismF (Ours)}                 & \textbf{39.62}         & \textbf{32.78} & \textbf{43.47} & \textbf{56.79} & \textbf{} & \textbf{40.27}        & \textbf{35.64} & \textbf{44.26} & 46.25 & \textbf{} & \textbf{16.98}        & \textbf{10.65} & \textbf{18.32} & \textbf{29.50} \\ \bottomrule
\end{tabular}}
\label{tab2}
\end{table*}

\subsubsection{Relation-Scaled Attention}
Relational semantics are incorporated into the fusion process through a relation-scaled attention mechanism. The relation-dependent temperature $\tau(\mathbf{r}_q)$ adaptively modulates the concentration of modality weights according to the specific semantic characteristics of different query relations:
\begin{equation}
\tau(\mathbf{r}_q) = \| \mathbf{r}_q \|_2 + \xi,
\end{equation}
\begin{equation}
\alpha_{m,q} = \frac{\exp(\widetilde{c}_{m,q} / \tau(\mathbf{r}_q))}{\sum_{m' \in \{s,v,t\}} \exp(\widetilde{c}_{m',q} / \tau(\mathbf{r}_q))},
\end{equation}
where $\xi$ is a small positive constant to avoid numerical instability.

Finally, the fused representation $\mathbf{e}_{fuse}$ is computed as:
\begin{equation}
{{\mathbf{e}}}_{fuse}=\sum_{m\in\left\{s,v,t\right\}}{\alpha_{m,q}\cdot{\mathbf{e}_{m,q}^{\text{final}}}}.
\end{equation}
The progressive nature of the fusion ensures that the new fused representation $\mathbf{e}_{fuse}$ captures both inter-modality interactions and modality-specific uncertainties, and we treat it as a new "modality" $\mathcal{F}$ (fuse).

\subsection{Link Prediction}
The overall training objective jointly optimizes the MuPE and PMF modules. Based on the enhanced modality-specific embeddings and fused representation, we compute the plausibility of a triple $(h, r, t)$ by employing a Tucker \citep{Tucker} score function for each modality $m\in\mathcal{M}\cup\left\{\mathcal{F}\right\}$, which is denoted as:
\begin{equation}
\begin{aligned}
    S_m\left(h, r, t^\prime\right) &= f_r\left(\mathbf{e}_m(h), \mathbf{e}_m(t^\prime)\right), \\
    S_m\left(h^\prime, r, t\right) &= f_r\left(\mathbf{e}_m(h^\prime), \mathbf{e}_m(t)\right),
\end{aligned}
\end{equation}
here, $f_r\left(\cdot,\cdot\right)$ is defined as:
\begin{equation}
f_r\left( \mathbf{e}_m(h),  \mathbf{e}_m(t)\right) = \langle \mathcal{W}, \mathbf{e}_m(h) \otimes r \otimes  \mathbf{e}_m(t) \rangle,
\end{equation}
where $\mathcal{W}$ is a core tensor, $\otimes$ denotes the tensor outer product, and
$\langle\cdot,\cdot\rangle$ denotes the Frobenius inner product.

By employing a unified negative log-likelihood loss, the training objective consolidates modality-specific and fused predictions, enabling the model to learn consistent scoring across modalities while exploiting complementary signals:
\begin{equation}
\begin{aligned}
\mathcal{L}_m = 
& - \sum_{m \in \mathcal{M} \cup \{\mathcal{F}\}}\sum_{\left(h, r, t\right) \in \mathcal{T}} \Bigg[
        \log \frac{\exp\left(S_m\left(h, r, t\right)\right)}
                  {\sum_{t^\prime \in \mathcal{E}} \exp\left(S_m\left(h, r, t^\prime\right)\right)} \\
& \quad + \log \frac{\exp\left(S_m\left(h, r, t\right)\right)}
                  {\sum_{h^\prime \in \mathcal{E}} \exp\left(S_m\left(h^\prime, r, t\right)\right)}
    \Bigg].
\end{aligned}
\end{equation}
The complete training objective integrates all components:
\begin{equation}
\mathcal{L}=\mathcal{L}_m+\gamma\mathcal{L}_{pd},
\end{equation}
where $\gamma$ controls the influence of the decoupling constraint during optimization and is tuned within a predefined range, as described in the Implementation Details section.

Through this unified optimization process, PrismF effectively aligns heterogeneous modality signals with relational semantics, resulting in more robust and discriminative representations while consistently improving overall performance in MMKGC tasks.

\begin{figure}
    \centering
    \includegraphics[width=1\linewidth]{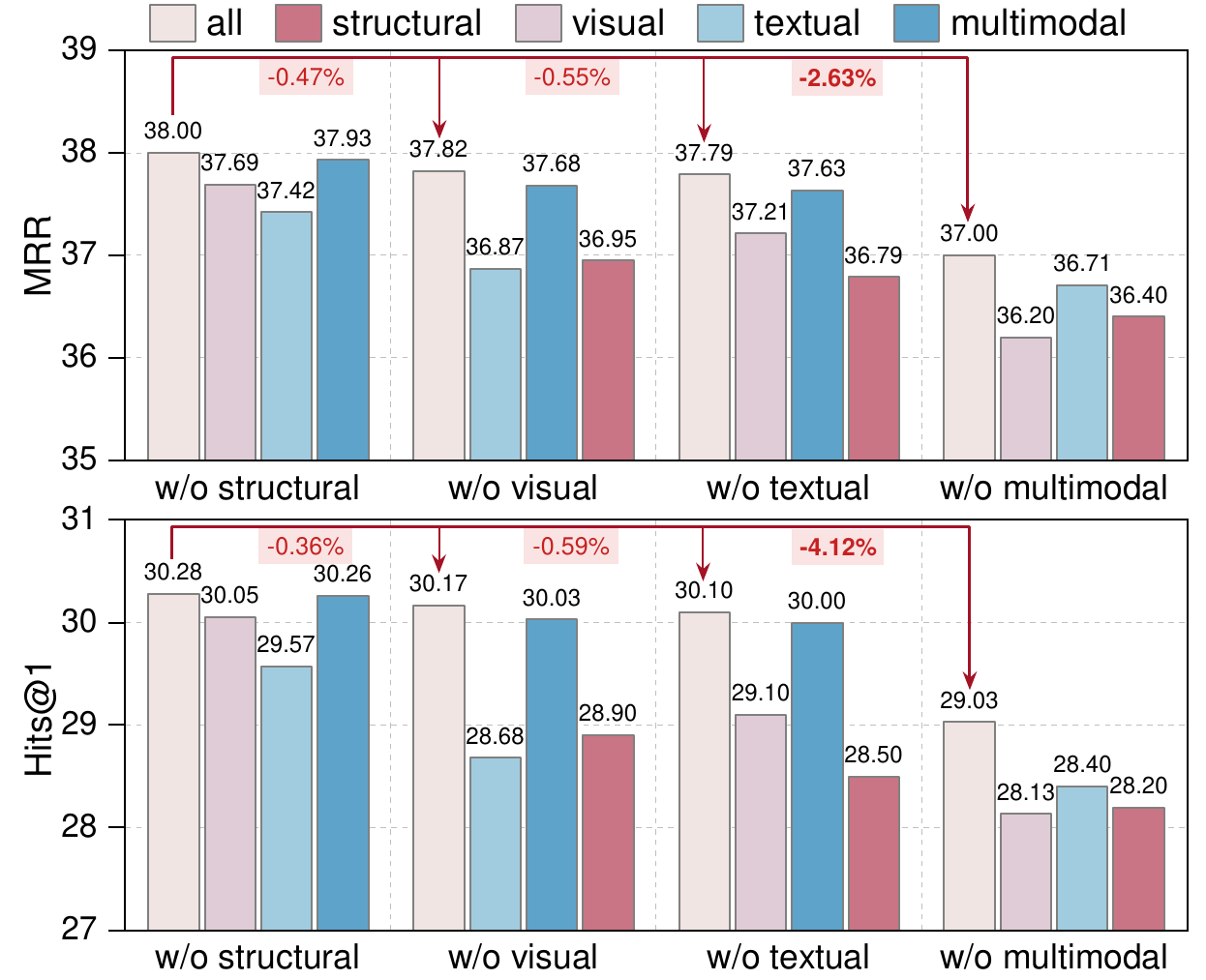}
    \caption{The MRR and Hits@1 results of removing each modality on DB15K.}
     
    \label{fig3}
\end{figure}

\section{Experiments}
\label{section4}
In this section, we evaluate our proposed PrismF through extensive experiments, aiming to address the following research questions:
\begin{itemize}
    \item \textbf{RQ1}: How does the performance of PrismF compare to the baseline models? (Section \ref{rq1})
    \item \textbf{RQ2}: How does different modality information impact the performance of PrismF? (Section \ref{rq2})    
    \item \textbf{RQ3}: How do the key modules affect the performance of PrismF? (Section \ref{rq3})
    \item \textbf{RQ4}: How do key hyperparameters influence PrismF’s performance? (Section \ref{rq4})
    \item \textbf{RQ5}: How robust is PrismF under challenging conditions such as modality missing and data sparsity? (Section \ref{rq5})
    \item \textbf{RQ6}: Are there qualitative case studies that intuitively demonstrate the effectiveness of PrismF? (Section \ref{rq7})
\end{itemize}

\subsection{Experimental Settings}
\subsubsection{Datasets}  In this paper, we employ three public MMKGC benchmarks, DB15K \citep{1}, MKG-Y \citep{MMRNS}, and KVC16K \citep{native}, to evaluate the model's performance. Each dataset comprises three modalities: structural triples, entity images, and textual descriptions. Table \ref{tab1} provides a statistical summary of these datasets.

\subsubsection{Evaluation Metrics}
We evaluate the MMKGC models with the link prediction task \citep{TransE}, a fundamental task of KGC. Following established works, we adopt rank-based metrics \citep{RotatE}, including Mean Reciprocal Rank (MRR) and Hits@K (K=1, 3, 10), to quantitatively measure model performance.

\subsubsection{Baseline}
To demonstrate the effectiveness of our model, we compare it with 17 baselines grouped into two groups: (1)Unimodal KGC models: TransE \citep{TransE}, DistMult \citep{Distmult}, ComplEx \citep{CompIEX}, RotatE \citep{RotatE}; (2) Multimodal KGC models: IKRL \citep{IKRL}, TransAE \citep{TransAE}, RSME \citep{RSME}, VBKGC \citep{vbkgc}, OTKGE \citep{OTKGE}, IMF \citep{IMF}, QEB \citep{QEB}, MANS \citep{MANS}, MMRNS \citep{MMRNS}, AdaMF \citep{ADAMF}, MYGO \citep{MYGO}, LBMKGC \citep{LBMKGC}, HFR-MKGC \citep{HFR-MKGC}.
\subsubsection{Implementation Details}

We implement our model in PyTorch and operate on an NVIDIA RTX 4090 GPU. During training, we tune the learning rate in \{$1\mathrm{e}^{-2}$, $1\mathrm{e}^{-3}$, $1\mathrm{e}^{-4}$\}, the dropout rate in \{0.2, 0.3, 0.4\}, embedding dimension in \{200, 300\}, the number of perspectives $N$ in \{2, 4, 6\}, the weight $\gamma$ in \{$1\mathrm{e}^{-2}$, $1\mathrm{e}^{-3}$, $1\mathrm{e}^{-4}$\}, and the batchsize is set to 1024. We fix the optimizer as Adam. For fairness, all baseline models are trained under the same experimental settings.

\begin{table}[!t]
\centering

\caption{Performance comparison of different settings. We explored the impact of each key component.}
\setlength{\tabcolsep}{1.3pt}   
\renewcommand{\arraystretch}{0.95} 
\begin{tabular}{lcccccc}
\toprule
 \multirow{2}{*}{\textbf{Setting}}& \multicolumn{2}{c}{\textbf{DB15K}}                                             & \multicolumn{2}{c}{\textbf{MKG-Y}}                                        & \multicolumn{2}{c}{\textbf{KVC16K}}                                  \\ \cmidrule{2-7} 
                                  & MRR  & Hits@1    & MRR  & Hits@1  & MRR  & Hits@1                         \\

\midrule

   w/o MuPE                         & 37.98  & 31.41  & 38.13  & 34.98  & 16.05    & 9.32 \\
   w/o loss $\mathcal{L}_{pd}$      & 38.18  & 31.64  & 38.31  & 35.17  & 16.27   & 9.61 \\
   w/o noise $\epsilon^i_m$         & 38.72  & 32.17 & 38.23  & 35.01  & 16.09   & 9.39 \\
\midrule

   w/o PMF                          & 37.91  & 31.22  & 37.36  & 33.96 & 16.32   & 9.78 \\
   w/ concat as $\mathbf{g}$                 & 38.03 & 30.62  & 37.98  & 34.82 & 16.58   & 10.12 \\
   w/o $\tau(\mathbf{r}_q)$         & 38.16  & 31.62 & 37.49  & 34.31  & 16.50  & 10.07 \\
\midrule
\multicolumn{1}{l}{\textbf{PrismF}} &\multicolumn{1}{c}{\textbf{39.62}} & \multicolumn{1}{c}{\textbf{32.78}}  &\multicolumn{1}{c}{\textbf{40.27}} & \multicolumn{1}{c}{\textbf{35.64}} &\multicolumn{1}{c}{\textbf{16.98}} & \multicolumn{1}{c}{\textbf{10.65}} \\
\bottomrule
\end{tabular}
\label{tab3}
\end{table}

\subsection{Performance Comparison (RQ1)} \label{rq1}
As shown in Table \ref{tab2}, PrismF achieves the best performance across all metrics. The gains are particularly evident on Hits@1, with relative improvements of 8.98\% on DB15K, 1.80\% on MKG-Y, and 11.17\% on KVC16K, indicating substantially better top-rank prediction. On the sparse and modality-imbalanced KVC16K dataset, PrismF achieves gains of 4.04\% in MRR, 6.14\% in Hits@3, and 3.26\% in Hits@10, suggesting that its improvements remain stable even when the multimodal signal is limited or uneven in quality.

These results reflect the overall design advantage of PrismF. Existing methods either rely primarily on structural patterns or employ multimodal fusion strategies that are insufficiently adaptive to noisy, ambiguous, or incomplete inputs. In contrast, PrismF simultaneously preserves fine-grained discriminative evidence and calibrates heterogeneous signals according to contextual reliability. This unified design allows the model to make more effective use of complementary modality information while avoiding over-reliance on unreliable cues, leading to stronger and more consistent reasoning performance across both balanced and challenging settings.

\begin{figure*}[t]
        \centering
        \includegraphics[width=0.95\linewidth]{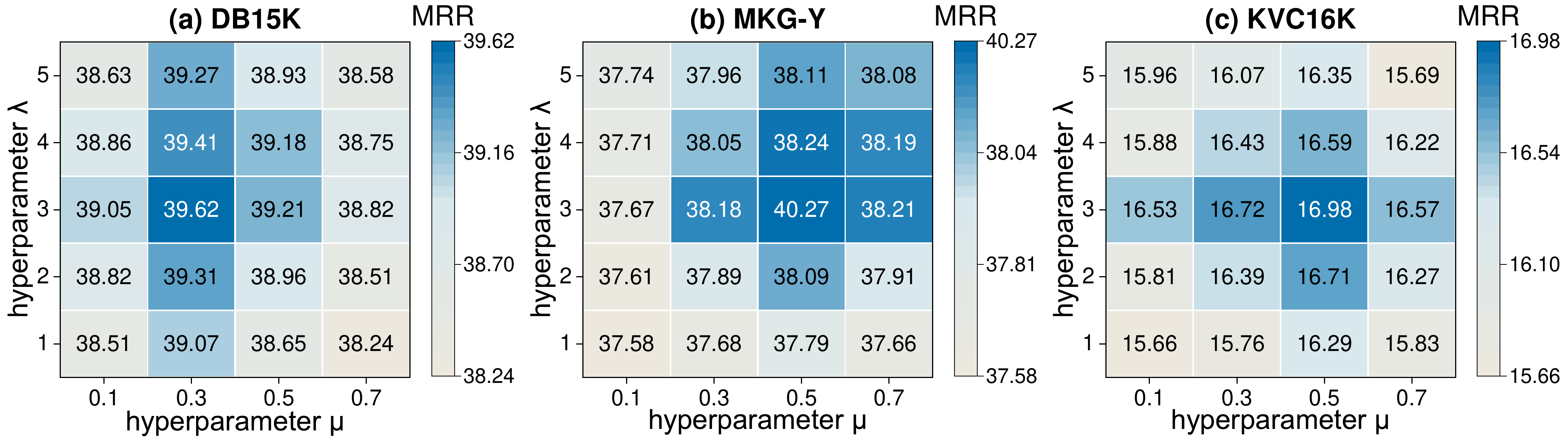}
        \caption{Hyperparameter performance analysis across all datasets.}
    \label{fig5}
\end{figure*}
\begin{figure}
        \centering
        \includegraphics[width=0.96\linewidth]{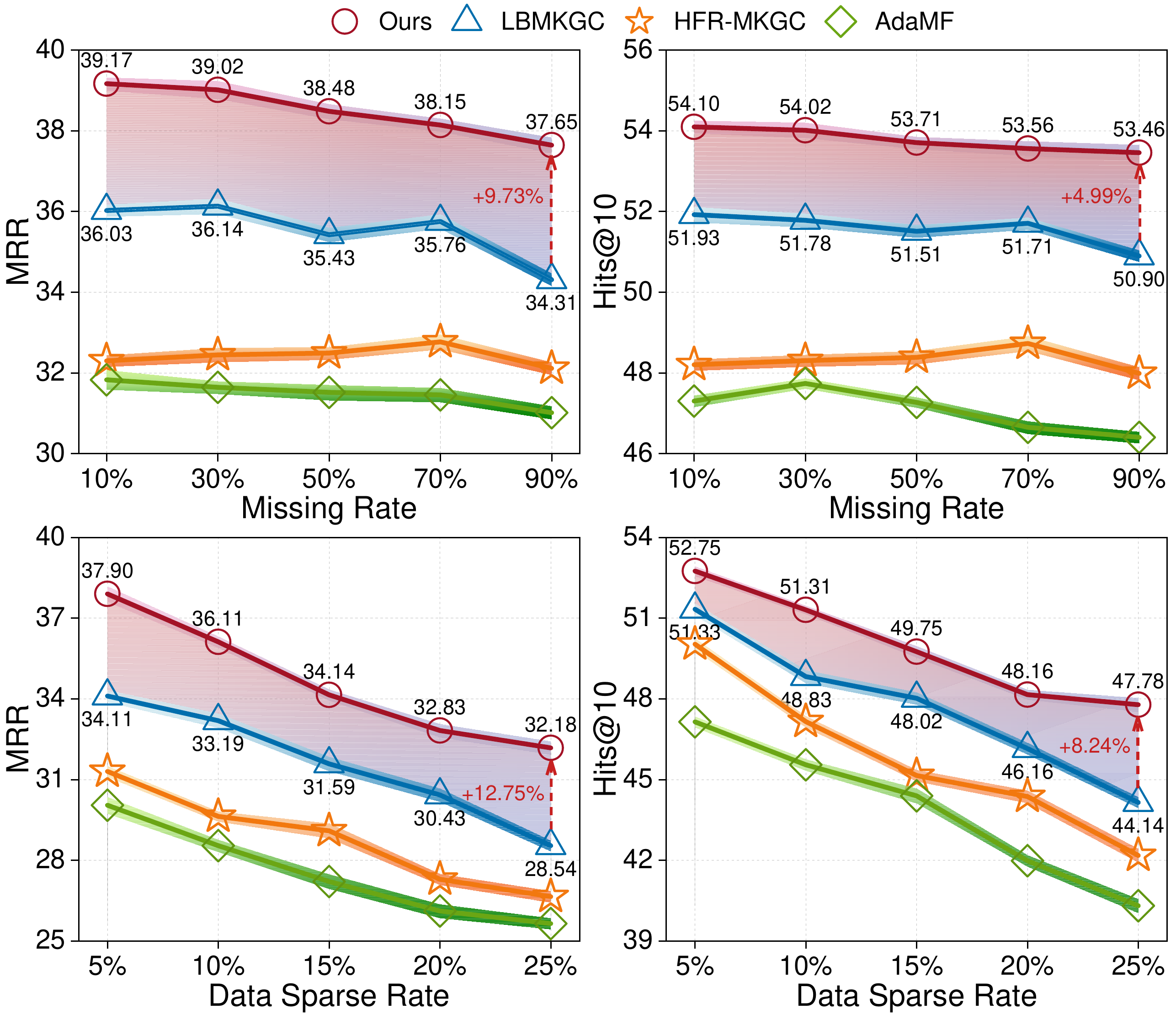}
        \caption{Performance comparison on the DB15K under two challenging scenarios: modality missing and data sparsity.}
    \label{fig4}
\end{figure}

\subsection{Effects of Modality (RQ2)}\label{rq2}
To evaluate the individual and joint contributions of different modalities, we conduct a modality ablation study by removing structural, visual, textual, or fused multimodal representations, where the fused representation is the output of the PMF module.

As shown in Figure~\ref{fig3}, removing any single modality consistently degrades performance, indicating that each modality provides complementary information for reasoning. Among all settings, removing the fused multimodal representation causes the largest performance drop, underscoring the crucial role of our architecture in integrating heterogeneous information. Through a progressive fusion of multi-perspective and cross-modal features, the fused representation not only aggregates modality-specific cues but also adaptively emphasizes the most informative signals under uncertainty. These results highlight that effective multimodal reasoning requires leveraging both the uniqueness and synergy of all modalities, rather than treating them equally and in isolation.

\subsection{Key Components (RQ3)} \label{rq3}
As shown in Table~\ref{tab3}, all ablated variants underperform the complete PrismF model, confirming that each component contributes to the overall performance, although their effects vary across datasets. On DB15K, removing MuPE and PMF reduces MRR by 4.14\% and 4.32\%, respectively. Replacing the mutual gating mechanism with feature concatenation results in the largest relative decrease in Hits@1 of 6.59\%, demonstrating its importance for accurate top-ranked prediction. On MKG-Y, removing PMF causes the largest MRR reduction of 7.23\%, followed by removing the relation-dependent temperature
term $\tau(\mathbf{r}_q)$, which leads to a 6.90\% decrease. These results indicate that progressive confidence calibration and relation-conditioned modality weighting are particularly important on this dataset. On KVC16K, removing MuPE and the stochastic refinement mechanism decreases MRR by 5.48\% and 5.24\%, respectively, highlighting the benefits of complementary
modality-specific representations and controlled perturbations. Overall, the results show that PrismF benefits from the coordinated interaction of its components rather than relying on any single module.

\subsection{Hyperparameter Analysis (RQ4)}
\label{rq4}
Figure~\ref{fig5} examines the effects of the two key hyperparameters in Equation~\ref{eq9}: the scaling factor $\lambda$, which controls the sharpness of the adjustment curve, and the threshold $\mu$, which determines its transition point. A small $\lambda$ produces an overly smooth adjustment, limiting the model's ability to differentiate modality contributions, whereas an excessively large $\lambda$ causes abrupt changes that may destabilize learning. Meanwhile, an inappropriate $\mu$ may trigger the adjustment either too early or too late, leading to suboptimal fusion. Overall, moderate settings provide a favorable balance between adaptive modality regulation and training stability, yielding consistently strong performance across datasets.

\begin{figure*}
        \centering
        \includegraphics[width=0.9\linewidth]{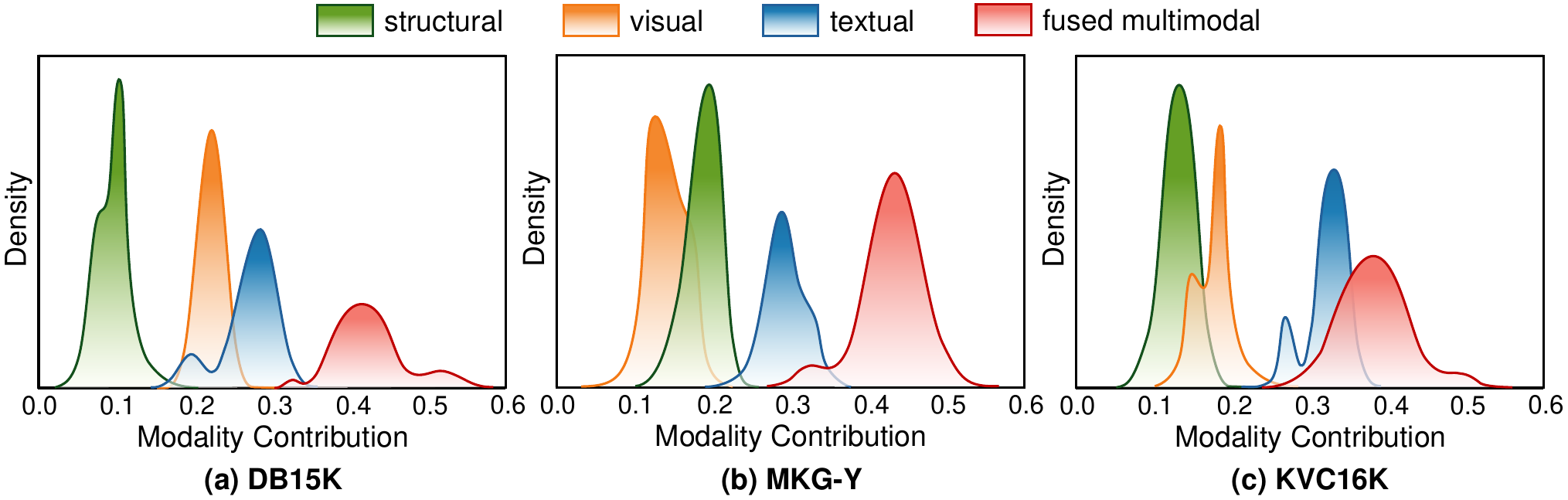}
        \caption{The visualization of modality contributions for PrismF. Here, we present the contributions of the structural, visual, textual, and fused multimodal during the reasoning process across all datasets.}
    \label{fig6}
\end{figure*}

\subsection{Complex Scenarios Analysis (RQ5)} \label{rq5}

To evaluate the robustness of PrismF, we conducted experiments under two challenging conditions: modality missing and data sparsity. The results are summarized in Figure \ref{fig4}.

\subsubsection{Modality Missing}
In this experiment, we follow prior work \citep{native} to simulate missing information by randomly discarding a portion of modality data. The results show that incomplete data significantly degrades MMKGC performance, but PrismF consistently outperforms all baselines. Interestingly, some baselines perform better at higher missing rates, likely due to their limited ability to effectively fuse multimodal signals. Removing certain modalities may reduce noise or conflicting signals, leading to unstable performance gains. In contrast, PrismF dynamically integrates available modalities and adjusts their contributions based on relevance and reliability, ensuring consistent reasoning with incomplete inputs. 

\subsubsection{Data Sparsity}
We simulate structural sparsity by progressively
removing 5\% to 25\% of the training triples. As shown in Figure~\ref{fig4}, all methods degrade
as structural supervision decreases, whereas PrismF maintains a more
gradual decline and consistently outperforms the baselines. At the highest sparsity rate of 25\%, PrismF achieves an MRR of 32.18 and a Hits@10 of 47.78, outperforming the strongest baseline by 12.75\% and 8.24\%, respectively. This result
suggests that MuPE preserves complementary modality-specific cues that can partially compensate for weakened structural evidence,
while PMF adaptively calibrates modality contributions when structural
signals are insufficient, thereby improving robustness under sparse
supervision.

\begin{figure}
        \centering
        \includegraphics[width=1\linewidth]{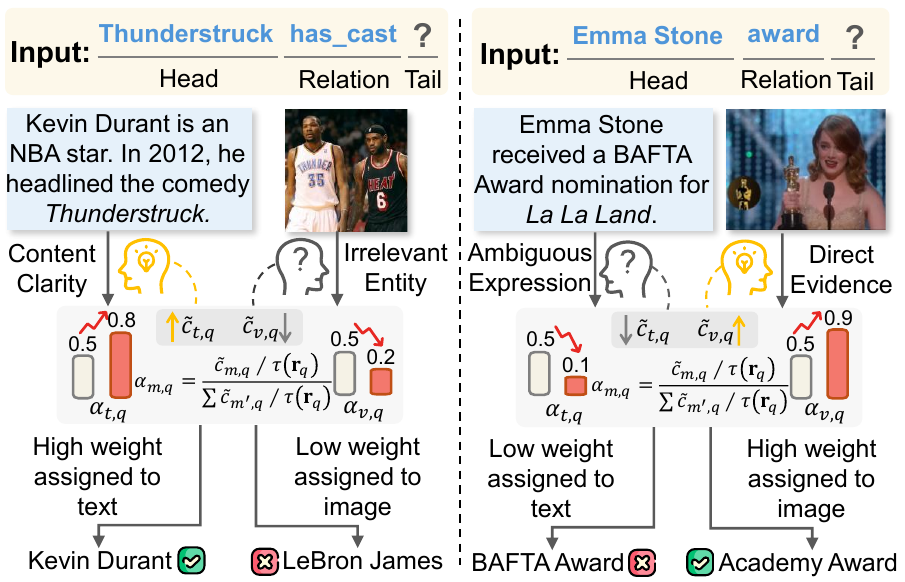}
        \caption{Case examples. PrismF assigns larger weights to the more informative modality based on content reliability, effectively mitigating noise and ambiguity.}
    \label{fig7}
\end{figure}

\subsection{Qualitative Analysis (RQ6)} \label{rq7}

\subsubsection{Robustness of Fusion Mechanism}

We further analyze the modality contribution distributions across all datasets to investigate the robustness of the proposed fusion mechanism. As shown in Figure~\ref{fig6}, PrismF maintains distinct contributions from structural, visual, textual, and fused representations, without any single modality exerting persistent dominance. The consistently substantial contribution of the fused representation indicates that complementary cross-modal evidence is effectively retained, while the distribution shifts across datasets suggest that the learned weighting adapts to differences in modality informativeness and reliability. This behavior is consistent with PMF, where the progressive factor limits premature reliance on concentrated confidence scores during early training, and the relation-dependent temperature regulates the concentration of modality weights under different query relations. Overall, these observations provide qualitative evidence that PrismF alleviates excessive modality dominance while preserving adaptive and reliability-aware multimodal fusion.

\subsubsection{Case Study}
To provide qualitative insight into the fusion behavior of PrismF,
we examine two illustrative queries from DB15K, as shown in
Figure~\ref{fig7}. For the query
(\textit{Thunderstruck}, \textit{has\_cast}, ?), the textual
description directly identifies \textit{Kevin Durant}, whereas the
visual input contains the irrelevant entity \textit{LeBron James}.
PrismF assigns a higher weight to the textual modality and a lower
weight to the visual modality, leading to the correct prediction.
For (\textit{Emma Stone}, \textit{award}, ?), the textual evidence
refers only to a BAFTA Award nomination, while the image provides
more direct evidence for \textit{Academy Award}. Accordingly, PrismF
places greater weight on the visual modality and produces the correct
answer. These examples illustrate how PrismF adjusts modality
contributions according to the informativeness of the available
evidence, thereby reducing the influence of ambiguous or irrelevant
signals.

\section{Conclusion}
In this paper, we propose PrismF, a unified framework for MMKGC. PrismF improves entity representations by jointly preserving fine-grained modality-specific cues and calibrating cross-modal interactions according to contextual reliability. Extensive experiments on DB15K, MKG-Y, and KVC16K demonstrate that PrismF achieves the strongest overall performance across the three benchmarks and remains effective under modality-missing and structurally sparse settings. These findings highlight that effective multimodal entity representation learning requires not only the integration of heterogeneous modalities, but also the preservation of discriminative intra-modal information and the reliability-aware calibration of cross-modal evidence. In future work, we plan to explore more general multimodal representation learning architectures and investigate how foundation models can be incorporated to further enhance knowledge-aware multimodal reasoning.

\begin{acks}
The authors sincerely thank the anonymous reviewers for their constructive comments, which helped improve this paper. This work was supported in part by the Natural Science Foundation of Hubei Province of China (No. 2025AFB653), the National Natural Science Foundation of China (No. 62207011, 62377009, 62507022), and the Open Fund of Hubei Key Laboratory of Big Data Intelligent Analysis and Application, Hubei University (No. 2025BDIAA01).
\end{acks}

\bibliographystyle{ACM-Reference-Format}
\bibliography{sample-base}


\end{document}